\documentclass[11pt,a4paper]{article}
\usepackage[hyperref]{emnlp-ijcnlp-2019}
\usepackage{times}
\usepackage{latexsym}

\usepackage{url}
\usepackage{helvet}  
\usepackage{courier}  
\usepackage{graphicx}  
\usepackage{multirow}
\usepackage{xspace}
\usepackage{amssymb}
\usepackage{amsmath}
\usepackage{subfigure}
\usepackage{booktabs}
\usepackage{bbm}
\usepackage{array}
\usepackage{bm}
\usepackage[noend]{algpseudocode}
\usepackage{algorithmicx,algorithm}
\usepackage{hyperref}

\DeclareMathOperator*{\argmax}{arg\,max}
\graphicspath{{figures/}} 

\aclfinalcopy 

\newcommand\bertbase{BERT$_{\textsc{BASE}}$\xspace}
\newcommand\bertlarge{BERT$_{\textsc{LARGE}}$\xspace}
\newcommand\nabertbase{NABERT$_{\textsc{BASE}}$\xspace}
\newcommand\nabertlarge{NABERT$_{\textsc{LARGE}}$\xspace}
\newcommand\mtmsnbase{MTMSN$_{\textsc{BASE}}$\xspace}
\newcommand\mtmsnlarge{MTMSN$_{\textsc{LARGE}}$\xspace}
\newcommand{\tabincell}[2]{\begin{tabular}{@{}#1@{}}#2\end{tabular}}

\title{A Multi-Type Multi-Span Network for \\ Reading Comprehension that Requires Discrete Reasoning}

\author{
Minghao Hu,
Yuxing Peng,
Zhen Huang,
Dongsheng Li \\ 
National University of Defense Technology, Changsha, China \\
{\tt \{huminghao09,pengyuxing,huangzhen,dsli\}@nudt.edu.cn}}

\date{}

\begin{document}
\maketitle

\begin{abstract}
Rapid progress has been made in the field of reading comprehension and question answering, where several systems have achieved human parity in some simplified settings.
However, the performance of these models degrades significantly when they are applied to more realistic scenarios, where answers are involved with various types, multiple text strings are correct answers, or discrete reasoning abilities are required.
In this paper, we introduce the Multi-Type Multi-Span Network (MTMSN), a neural reading comprehension model that combines a multi-type answer predictor designed to support various answer types (e.g., span, count, negation, and arithmetic expression) with a multi-span extraction method for dynamically producing one or multiple text spans.
In addition, an arithmetic expression reranking mechanism is proposed to rank expression candidates for further confirming the prediction.
Experiments show that our model achieves 79.9 F1 on the DROP hidden test set, creating new state-of-the-art results.
Source code\footnote{\url{https://github.com/huminghao16/MTMSN}} is released to facilitate future work.
\end{abstract}
\section{Introduction}

This paper considers the reading comprehension task in which some \emph{discrete-reasoning} abilities are needed to correctly answer questions.
Specifically, we focus on a new reading comprehension dataset called DROP~\cite{dua2019drop}, which requires Discrete Reasoning Over the content of Paragraphs to obtain the final answer.
Unlike previous benchmarks such as CNN/DM~\cite{hermann2015teaching} and SQuAD~\cite{Rajpurkar16} that have been well solved~\cite{chen2016thorough,devlin2018bert}, DROP is substantially more challenging in three ways.
First, the answers to the questions involve a wide range of types such as numbers, dates, or text strings.
Therefore, various kinds of prediction strategies are required to successfully find the answers.  
Second, rather than restricting the answer to be a span of text, DROP loosens the constraint so that answers may be a set of multiple text strings.
Third, for questions that require discrete reasoning, a system must have a more comprehensive understanding of the context and be able to perform numerical operations such as addition, counting, or sorting. 

Existing approaches, when applied to this more realistic scenario, have three problems.
First, to produce various answer types, \citet{dua2019drop} extend previous one-type answer prediction~\cite{seo2016bidirectional} to multi-type prediction that supports span extraction, counting, and addition/subtraction.
However, they have not fully considered all potential types. 
Take the question ``\emph{What percent are not non-families?}'' and the passage snippet ``\emph{39.9\% were non-families}'' as an example, a \emph{negation} operation is required to infer the answer.
Second, previous reading comprehension models~\cite{wang2017gated,yu2018fast,hu2017reinforced} are designed to produce one single span as the answer. 
But for some questions such as ``\emph{Which ancestral groups are smaller than 11\%?}'', there may exist several spans as correct answers (e.g., ``\emph{Italian}'', ``\emph{English}'', and ``\emph{Polish}''), which can not be well handled by these works.
Third, to support numerical reasoning, prior work~\cite{dua2019drop} learns to predict signed numbers for obtaining an arithmetic expression that can be executed by a symbolic system. 
Nevertheless, the prediction of each signed number is isolated, and the expression's context information has not been considered. As a result, obviously-wrong expressions, such as all predicted signs are either minus or zero, are likely produced.

To address the above issues, we introduce the Multi-Type Multi-Span Network (MTMSN), a neural reading comprehension model for predicting various types of answers as well as dynamically extracting one or multiple spans. 
MTMSN utilizes a series of pre-trained Transformer blocks~\cite{devlin2018bert} to obtain a deep bidirectional context representation.
On top of it, a multi-type answer predictor is proposed to not only support previous prediction strategies such as span, count number, and arithmetic expression, but also add a new type of logical negation.
This results in a wider range of coverage of answer types, which turns out to be crucial to performance.
Besides, rather than always producing one single span, we present a multi-span extraction method to produce multiple answers. 
The model first predicts the number of answers, and then extracts non-overlapped spans to the specific amount.
In this way, the model can learn to dynamically extract one or multiple spans, thus being beneficial for multi-answer cases.
In addition, we propose an arithmetic expression reranking mechanism to rank expression candidates that are decoded by beam search, so that their context information can be considered during reranking to further confirm the prediction.

Our MTMSN model outperforms all existing approaches on the DROP hidden test set by achieving 79.9 F1 score, a 32.9\% absolute gain over prior best work at the time of submission. 
To make a fair comparison, we also construct a baseline that uses the same BERT-based encoder.
Again, MTMSN surpasses it by obtaining a 13.2 F1 increase on the development set.
We also provide an in-depth ablation study to show the effectiveness of our proposed methods, analyze performance breakdown by different answer types, and give some qualitative examples as well as error analysis.
\section{Task Description}
In the reading comprehension task that requires discrete reasoning, a passage and a question are given.
The goal is to predict an answer to the question by reading and understanding the passage.
Unlike previous dataset such as SQuAD~\cite{Rajpurkar16} where the answer is limited to be a single span of text, DROP loosens the constraint so that the answer involves various types such as number, date, or span of text (Figure \ref{fig:drop-example}). 
Moreover, the answer can be multiple text strings instead of single continuous span ($\mathbf{A}_2$).
To successfully find the answer, some discrete reasoning abilities, such as sorting ($\mathbf{A}_1$), subtraction ($\mathbf{A}_3$), and negation ($\mathbf{A}_4$), are required.

\begin{figure}[t]
\center
\fbox{\parbox{0.95\columnwidth}{
\begin{small}
$\mathbf{Passage}$: As of the census of 2000, there were 218,590 people, 79,667 households, ... 22.5\% were of German people, 13.1\% Irish people, 9.8\% Italian people, ...

$\mathbf{Q}_1$: Which group from the census is larger: German or Irish?

$\mathbf{A}_1$: German

$\mathbf{Q}_2$: Which ancestral groups are at least 10\%?

$\mathbf{A}_2$: German, Irish

$\mathbf{Q}_3$: How many more people are there than households?

$\mathbf{A}_3$: 138,923

$\mathbf{Q}_4$: How many percent were not German?

$\mathbf{A}_4$: 77.5

\end{small}
}}
\caption{Question-answer pairs along with a passage from the DROP dataset.}
\label{fig:drop-example}
\end{figure}

\section{Our Approach}
\begin{figure*}
\center
\includegraphics[width=0.95\textwidth]{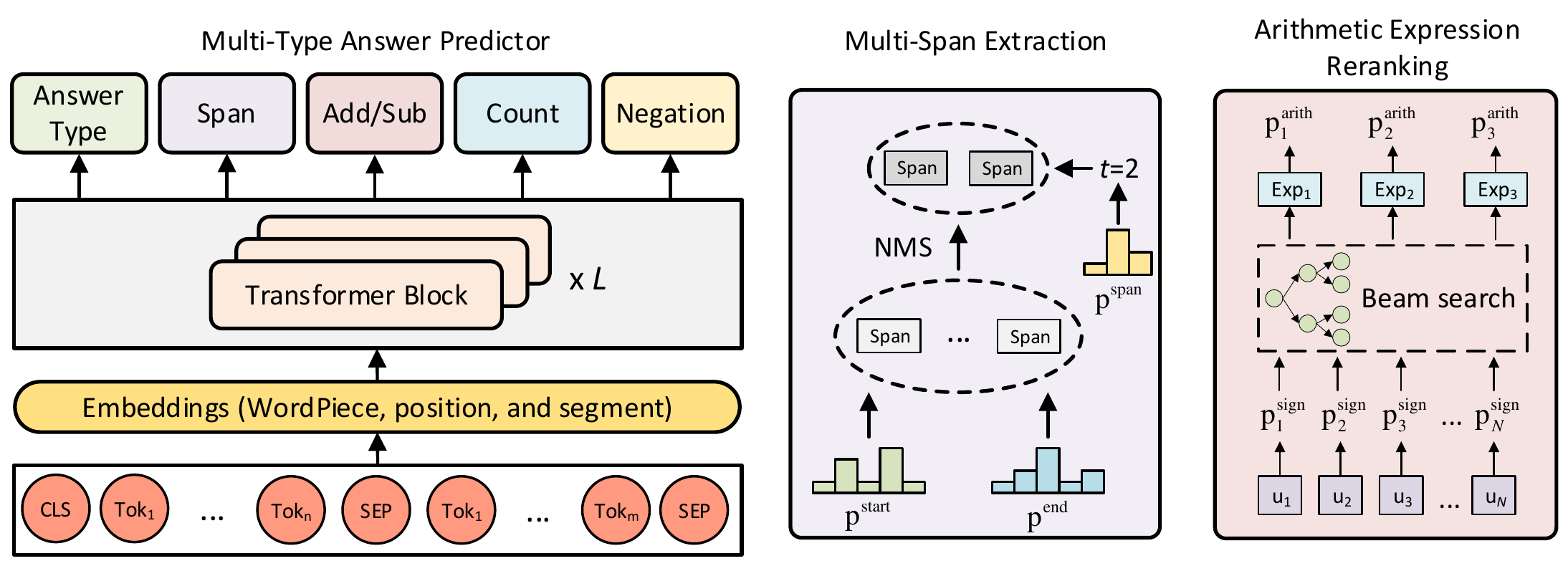}
\caption{An illustration of MTMSN architecture. 
The multi-type answer predictor supports four kinds of answer types including span, addition/subtraction, count, and negation.
A multi-span extraction method is proposed to dynamically produce one or several spans.
The arithmetic expression reranking mechanism aims to rank expression candidates that are decoded by beam search for further validating the prediction.}
\label{fig:mtmsn-overview}
\end{figure*}

Figure \ref{fig:mtmsn-overview} gives an overview of our model that aims to combine neural reading comprehension with numerical reasoning.
Our model uses BERT~\cite{devlin2018bert} as encoder: we map word embeddings into contextualized representations using pre-trained Transformer blocks~\cite{vaswani2017attention} (\S\ref{sec:bert-enc}).
Based on the representations, we employ a multi-type answer predictor that is able to produce four answer types: (1) span from the text; (2) arithmetic expression; (3) count number; (4) negation on numbers (\S\ref{sec:ans-pred}).
Following \citet{dua2019drop}, we first predict the answer type of a given passage-question pair, and then adopt individual prediction strategies.
To support multi-span extraction (\S\ref{sec:multi-span}), the model explicitly predicts the number of answer spans. It then outputs non-overlapped spans until the specific amount is reached.
Moreover, we do not directly use the arithmetic expression that possesses the maximum probability, but instead re-rank several expression candidates that are decoded by beam search to further confirm the prediction (\S\ref{sec:exp-rerank}).
Finally, the model is trained under weakly-supervised signals to maximize the marginal likelihood over all possible annotations (\S\ref{sec:train-infer}).

\subsection{BERT-Based Encoder 	\label{sec:bert-enc}}
To obtain a universal representation for both the question and the passage, we utilize BERT~\cite{devlin2018bert}, a pre-trained deep bidirectional Transformer model that achieves state-of-the-art performance across various tasks, as the encoder.

Specifically, we first tokenize the question and the passage using the WordPiece vocabulary~\cite{wu2016google}, and then generate the input sequence by concatenating a \texttt{[CLS]} token, the tokenized question, a \texttt{[SEP]} token, the tokenized passage, and a final \texttt{[SEP]} token.
For each token in the sequence, its input representation is the element-wise addition of WordPiece embeddings, positional embeddings, and segment embeddings~\cite{devlin2018bert}. 
As a result, a list of input embeddings $\mathbf{H}_0 \in\mathbb{R}^{T \times D}$ can be obtained, where $D$ is the hidden size and $T$ is the sequence length. A series of $L$ pre-trained Transformer blocks are then used to project the input embeddings into contextualized representations $\mathbf{H}_i$ as:
\begin{equation}
	\mathbf{H}_i = \mathrm{TransformerBlock}(\mathbf{H}_{i-1}), \forall i \in [1,L] \nonumber
\end{equation}
Here, we omit a detailed introduction of the block architecture and refer readers to \citet{vaswani2017attention} for more details.

\subsection{Multi-Type Answer Predictor	\label{sec:ans-pred}}
Rather than restricting the answer to always be a span of text, the discrete-reasoning reading comprehension task involves different answer types (e.g., number, date, span of text).
Following \citet{dua2019drop}, we design a multi-type answer predictor to selectively produce different kinds of answers such as span, count number, and arithmetic expression.
To further increase answer coverage, we propose adding a new answer type to support logical negation.
Moreover, unlike prior work that separately predicts passage spans and question spans, our approach directly extracts spans from the input sequence.

\paragraph{Answer type prediction}
Inspired by the Augmented QANet model~\cite{dua2019drop}, we use the contextualized token representations from the last four blocks ($\mathbf{H}_{L-3}$, ..., $\mathbf{H}_L$) as the inputs to our answer predictor, which are denoted as $\mathbf{M}_0$, $\mathbf{M}_1$, $\mathbf{M}_2$, $\mathbf{M}_3$, respectively.
To predict the answer type, we first split the representation $\mathbf{M}_2$ into a question representation $\mathbf{Q}_2$ and a passage representation $\mathbf{P}_2$ according to the index of intermediate \texttt{[SEP]} token.
Then the model computes two vectors $\mathbf{h}^{\mathbf{Q}_2}$ and $\mathbf{h}^{\mathbf{P}_2}$ that summarize the question and passage information respectively:
\begin{equation}
	\boldsymbol{\alpha}^Q = \mathrm{softmax}(\mathbf{W}^Q \mathbf{Q}_2), \quad \mathbf{h}^{\mathbf{Q}_2} = \boldsymbol{\alpha}^Q \mathbf{Q}_2 \nonumber
\end{equation}
where $\mathbf{h}^{\mathbf{P}_2}$ is computed in a similar way over $\mathbf{P}_2$.

Next, we calculate a probability distribution to represent the choices of different answer types as:
\begin{equation}
	\mathbf{p}^{\mathrm{type}} = \mathrm{softmax}(\mathrm{FFN}([\mathbf{h}^{\mathbf{Q}_2}; \mathbf{h}^{\mathbf{P}_2}; \mathbf{h}^{\mathrm{CLS}}])) \nonumber
\end{equation}
Here, $\mathbf{h}^{\mathrm{CLS}}$ is the first vector in the final contextualized representation $\mathbf{M}_3$, and $\mathrm{FFN}$ denotes a feed-forward network consisting of two linear projections with a GeLU activation~\cite{hendrycks2016bridging} followed by a layer normalization~\cite{lei2016layer} in between.

\paragraph{Span}
To extract the answer either from the passage or from the question, we combine the gating mechanism of \citet{wang2017gated} with the standard decoding strategy of \citet{seo2016bidirectional} to predict the starting and ending positions across the entire sequence.
Specifically, we first compute three vectors, namely $\mathbf{g}^{\mathbf{Q}_0}$, $\mathbf{g}^{\mathbf{Q}_1}$, $\mathbf{g}^{\mathbf{Q}_2}$, that summarize the question information among different levels of question representations:
\begin{equation}
	\boldsymbol{\beta}^Q = \mathrm{softmax}(\mathrm{FFN}(\mathbf{Q}_2), \quad \mathbf{g}^{\mathbf{Q}_2} = \boldsymbol{\beta}^Q \mathbf{Q}_2 \nonumber
\end{equation}
where $\mathbf{g}^{\mathbf{Q}_0}$ and $\mathbf{g}^{\mathbf{Q}_1}$ are computed over $\mathbf{Q}_0$ and $\mathbf{Q}_1$ respectively, in a similar way as described above.

Then we compute the probabilities of the starting and ending indices of the answer span from the input sequence as:
\begin{align}
	\bar{\mathbf{M}}^\mathrm{start} &= [\mathbf{M}_2;\mathbf{M}_0; \mathbf{g}^{\mathbf{Q}_2} \otimes \mathbf{M}_2; \mathbf{g}^{\mathbf{Q}_0} \otimes \mathbf{M}_0], \nonumber \\
	\bar{\mathbf{M}}^\mathrm{end} &= [\mathbf{M}_2;\mathbf{M}_1; \mathbf{g}^{\mathbf{Q}_2} \otimes \mathbf{M}_2; \mathbf{g}^{\mathbf{Q}_1} \otimes \mathbf{M}_1], \nonumber \\ 
    \mathbf{p}^\mathrm{start} &= \mathrm{softmax}(\mathbf{W}^S \bar{\mathbf{M}}^\mathrm{start}), \nonumber \\ 
    \mathbf{p}^\mathrm{end} &= \mathrm{softmax}(\mathbf{W}^E \bar{\mathbf{M}}^\mathrm{end}) \nonumber
\end{align}
where $\otimes$ denotes the outer product between the vector $\mathbf{g}$ and each token representation in $\mathbf{M}$.

\paragraph{Arithmetic expression} 
In order to model the process of performing addition or subtraction among multiple numbers mentioned in the passage, we assign a three-way categorical variable (plus, minus, or zero) for each number to indicate its sign, similar to \citet{dua2019drop}. 
As a result, an arithmetic expression that has a number as the final answer can be obtained and easily evaluated.

Specifically, for each number mentioned in the passage, we gather its corresponding representation from the concatenation of $\mathbf{M}_2$ and $\mathbf{M}_3$, eventually yielding $\mathbf{U} = (\mathbf{u}_1, ..., \mathbf{u}_N) \in\mathbb{R}^{N \times 2*D}$ where $N$ numbers exist. 
Then the probabilities of the $i$-th number being assigned a plus, minus or zero is computed as:
\begin{equation}
	\mathbf{p}^{\mathrm{sign}}_i = \mathrm{softmax}(\mathrm{FFN}([\mathbf{u}_i; \mathbf{h}^{\mathbf{Q}_2}; \mathbf{h}^{\mathbf{P}_2}; \mathbf{h}^{\mathrm{CLS}}])) \nonumber
\end{equation}

\paragraph{Count} 
We consider the ability of counting entities and model it as a multi-class classification problem. 
To achieve this, the model first produces a vector $\mathbf{h}^{\mathbf{U}}$ that summarizes the important information among all mentioned numbers, and then computes a counting probability distribution as:
\begin{align}
	\boldsymbol{\alpha}^U &= \mathrm{softmax}(\mathbf{W}^U \mathbf{U}), \quad \mathbf{h}^{\mathbf{U}} = \boldsymbol{\alpha}^U \mathbf{U} \nonumber, \\
	\mathbf{p}^{\mathrm{count}} &= \mathrm{softmax}(\mathrm{FFN}([\mathbf{h}^{\mathbf{U}}; \mathbf{h}^{\mathbf{Q}_2}; \mathbf{h}^{\mathbf{P}_2}; \mathbf{h}^{\mathrm{CLS}}])) \nonumber
\end{align}

\paragraph{Negation}
One obvious but important linguistic phenomenon that prior work fails to capture is \emph{negation}. 
We find there are many cases in DROP that require to perform logical negation on numbers.
The question ($\mathbf{Q}_4$) in Figure \ref{fig:drop-example} gives a qualitative example of this phenomenon.
To model this phenomenon, we assign a two-way categorical variable for each number to indicate whether a negation operation should be performed.
Then we compute the probabilities of logical negation on the $i$-th number as:
\begin{equation}
	\mathbf{p}^{\mathrm{negation}}_i = \mathrm{softmax}(\mathrm{FFN}([\mathbf{u}_i; \mathbf{h}^{\mathbf{Q}_2}; \mathbf{h}^{\mathbf{P}_2}; \mathbf{h}^{\mathrm{CLS}}])) \nonumber
\end{equation}

\subsection{Multi-Span Extraction	\label{sec:multi-span}}
Although existing reading comprehension tasks focus exclusively on finding one span of text as the final answer, DROP loosens the restriction so that the answer to the question may be several text spans. 
Therefore, specific adaption should be made to extend previous single-span extraction to multi-span scenario.

To do this, we propose directly predicting the number of spans and model it as a classification problem. 
This is achieved by computing a probability distribution on span amount as
\begin{equation}
	\mathbf{p}^{\mathrm{span}} = \mathrm{softmax}(\mathrm{FFN}([\mathbf{h}^{\mathbf{Q}_2}; \mathbf{h}^{\mathbf{P}_2}; \mathbf{h}^{\mathrm{CLS}}])) \nonumber
\end{equation}

To extract non-overlapped spans to the specific amount, we adopt the non-maximum suppression (NMS) algorithm~\cite{rosenfeld1971edge} that is widely used in computer vision for pruning redundant bounding boxes, as shown in Algorithm \ref{algo:nms}.
Concretely, the model first proposes a set of top-$K$ spans $\mathbf{S}$ according to the descending order of the span score, which is computed as $\mathbf{p}^\mathrm{start}_k \mathbf{p}^\mathrm{end}_l$ for the span $(k, l)$.
It also predicts the amount of extracted spans $t$ from $\mathbf{p}^{\mathrm{span}}$, and initializes a new set $\tilde{\mathbf{S}}$.
Next, we add the span $\mathbf{s}_i$ that possesses the maximum span score to the set $\tilde{\mathbf{S}}$, and remove it from $\mathbf{S}$. 
We also delete any remaining span $\mathbf{s}_j$ that overlaps with $\mathbf{s}_i$, where the degree of overlap is measured using the text-level F1 function. 
This process is repeated for remaining spans in $\mathbf{S}$, until $\mathbf{S}$ is empty or the size of $\tilde{\mathbf{S}}$ reaches $t$. 

\begin{algorithm}[h]
\small
\caption{Multi-span extraction} 
\label{algo:nms}
{\bf Input:} 
$\mathbf{p}^\mathrm{start}$; $\mathbf{p}^\mathrm{end}$; $\mathbf{p}^{\mathrm{span}}$
\begin{algorithmic}[1]
\State Generate the set $\mathbf{S}$ by extracting top-$K$ spans
\State Sort $\mathbf{S}$ in descending order of span scores
\State $t = \argmax \mathbf{p}^{\mathrm{span}} + 1$
\State Initialize $\tilde{\mathbf{S}} = \{\}$ 
\While{$\mathbf{S} \ne \{\}$ and $|\tilde{\mathbf{S}}| < t$} 
	\For{$\mathbf{s}_i$ in $\mathbf{S}$}
		\State Add span $\mathbf{s}_i$ to $\tilde{\mathbf{S}}$
		\State Remove span $\mathbf{s}_i$ from $\mathbf{S}$
		\For{$\mathbf{s}_j$ in $\mathbf{S}$}
			\If{$\mathrm{f1}(\mathbf{s}_i, \mathbf{s}_j) > 0$}
				\State Remove span $\mathbf{s}_j$ from $\mathbf{S}$
			\EndIf
		\EndFor
	\EndFor
\EndWhile
\State \Return $\tilde{\mathbf{S}}$
\end{algorithmic}
\end{algorithm}

\subsection{Arithmetic Expression Reranking	\label{sec:exp-rerank}}
As discussed in \S\ref{sec:ans-pred}, we model the phenomenon of discrete reasoning on numbers by learning to predict a plus, minus, or zero for each number in the passage.
In this way, an arithmetic expression composed of signed numbers can be obtained, where the final answer can be deduced by performing simple arithmetic computation. 
However, since the sign of each number is only determined by the number representation and some coarse-grained global representations, the context information of the expression itself has not been considered.
As a result, the model may predict some obviously wrong expressions (e.g., the signs that have maximum probabilities are either minus or zero, resulting in a large negative value).
Therefore, in order to further validate the prediction, it is necessary to rank several highly confident expression candidates using the representation summarized from the expression's context.

Specifically, we use beam search to produce top-ranked arithmetic expressions, which are sent back to the network for reranking.
Since each expression consists of several signed numbers, we construct an expression representation by taking both the numbers and the signs into account.
For each number in the expression, we gather its corresponding vector from the representation $\mathbf{U}$.
As for the signs, we initialize an embedding matrix $\mathbf{E} \in\mathbb{R}^{3 \times 2*D}$, and find the sign embeddings for each signed number. 
In this way, given the $i$-th expression that contains $M$ signed numbers at most, we can obtain number vectors $\mathbf{V}_i \in\mathbb{R}^{M \times 2*D}$ as well as sign embeddings $\mathbf{C}_i \in\mathbb{R}^{M \times 2*D}$.
Then the expression representation along with the reranking probability can be calculated as:
\begin{align}
	\boldsymbol{\alpha}^V_i &= \mathrm{softmax}(\mathbf{W}^V (\mathbf{V}_i + \mathbf{C}_i))\nonumber, \\
	\mathbf{h}^{\mathbf{V}}_i &= \boldsymbol{\alpha}^V_i (\mathbf{V}_i + \mathbf{C}_i) \nonumber, \\
	\mathbf{p}^{\mathrm{arith}}_i &= \mathrm{softmax}(\mathrm{FFN}([\mathbf{h}^{\mathbf{V}}_i; \mathbf{h}^{\mathbf{Q}_2}; \mathbf{h}^{\mathbf{P}_2}; \mathbf{h}^{\mathrm{CLS}}])) \nonumber
\end{align}

\subsection{Training and Inference	\label{sec:train-infer}}
Since DROP does not indicate the answer type but only provides the answer string, we therefore adopt the weakly supervised annotation scheme, as suggested in \citet{berant2013semantic,dua2019drop}. 
We find all possible annotations that point to the gold answer, including matching spans, arithmetic expressions, correct count numbers, negation operations, and the number of spans.
We use simple rules to search over all mentioned numbers to find potential negations. That is, if 100 minus a number is equal to the answer, then a negation occurs on this number.
Besides, we only search the addition/subtraction of three numbers at most due to the exponential search space.

To train our model, we propose using a two-step training method composed of an inference step and a training step.
In the first step, we use the model to predict the probabilities of sign assignments for numbers. 
If there exists any annotation of arithmetic expressions, we run beam search to produce expression candidates and label them as either correct or wrong, which are later used for supervising the reranking component.
In the second step, we adopt the marginal likelihood objective function~\cite{clark2017simple}, which sums over the probabilities of all possible annotations including the above labeled expressions.
Notice that there are two objective functions for the multi-span component: one is a distantly-supervised loss that maximizes the probabilities of all matching spans, and the other is a classification loss that maximizes the probability on span amount.

At test time, the model first chooses the answer type and then performs specific prediction strategies. 
For the span type, we use Algorithm \ref{algo:nms} for decoding.
If the type is addition/subtraction, arithmetic expression candidates will be proposed and further reranked.
The expression with the maximum product of cumulative sign probability and reranking probability is chosen.
As for the counting type, we choose the number that has the maximum counting probability.
Finally, if the type is negation, we find the number that possesses the largest negation probability, and then output the answer as 100 minus this number.

\section{Experiments}

\begin{table*}
\begin{center}
\begin{tabular}{l|cccc}
\toprule
\multirow{2}*{ Model } & \multicolumn{2}{c}{ Dev } & \multicolumn{2}{c}{ Test } \\
 & EM & F1 & EM & F1 \\ 
\midrule
\midrule
Heuristic Baseline~\cite{dua2019drop}               & 4.28 & 8.07 & 4.18 & 8.59 \\ 
Semantic Role Labeling~\cite{carreras2004introduction}            	   & 11.03 & 13.67 & 10.87 & 13.35 \\
BiDAF~\cite{seo2016bidirectional}				   & 26.06 & 28.85 & 24.75 & 27.49 \\
QANet+ELMo~\cite{yu2018fast}			   & 27.71 & 30.33 & 27.08 & 29.67 \\
\bertbase~\cite{devlin2018bert}			   & 30.10 & 33.36 & 29.45 & 32.70 \\
NAQANet~\cite{dua2019drop}				   & 46.20 & 49.24 & 44.07 & 47.01 \\
\midrule
\nabertbase				   & 55.82 & 58.75 & - & - \\
\nabertlarge			   & 64.61 & 67.35 & - & - \\
\mtmsnbase                 & 68.17 & 72.81 & - & - \\
\mtmsnlarge                & \textbf{76.68} & \textbf{80.54} & \textbf{75.85} & \textbf{79.88} \\
\midrule
Human Performance~\cite{dua2019drop}	& - & - & 92.38 & 95.98 \\
\bottomrule
\end{tabular}
\caption{\label{table:drop-result} The performance of MTMSN and other competing approaches on DROP dev and test set.}
\end{center}
\end{table*}

\subsection{Implementation Details}
\paragraph{Dataset}
We consider the reading comprehension benchmark that requires Discrete Reasoning Over Paragraphs (DROP)~\cite{dua2019drop} to train and evaluate our model.
DROP contains crowdsourced, adversarially-created, 96.6K question-answer pairs, with 77.4K for training, 9.5K for validation, and another 9.6K hidden examples for testing.
Passages are extracted from Wikipedia articles and the answer to each question involves various types such as number, date, or text string.
Some answers may even be a set of multiple spans of text in the passage.
To find the answers, a comprehensive understanding of the context as well as the ability of numerical reasoning are required.

\begin{table}
	\begin{center}
		\begin{tabular}{l|cccc}
			\toprule
			\multirow{2}*{ Model }  & \multicolumn{2}{c}{ BASE } & \multicolumn{2}{c}{ LARGE } \\
			& EM & F1 & EM & F1 \\ 
			\midrule
			MTMSN                & 68.2 & 72.8 & 76.7 & 80.5 \\
			w/o Add/Sub            & 46.7 & 51.3 & 53.8 & 58.0 \\
			w/o Count              & 62.5 & 66.4 & 71.8 & 75.6 \\
			w/o Negation           & 59.4 & 63.6 & 67.2 & 70.9 \\
			w/o Multi-Span	     & 67.5 & 70.7 & 75.6 & 78.4 \\
			w/o Reranking          & 66.9 & 71.2 & 74.9 & 78.7 \\
			\bottomrule
		\end{tabular}
		\caption{\label{table:ablation1} Ablation tests of base and large models on the DROP dev set.}
	\end{center}
\end{table}

\paragraph{Model settings}
We build our model upon two publicly available uncased versions of BERT: \bertbase and \bertlarge\footnote{\bertbase is the original version while \bertlarge is the model augmented with n-gram masking and synthetic self-training: \url{https://github.com/google-research/bert.}}, and refer readers to~\citet{devlin2018bert} for details on model sizes.
We use Adam optimizer with a learning rate of 3e-5 and warmup over the first 5\% steps to train.
The maximum number of epochs is set to 10 for base models and 5 for large models, while the batch size is 12 or 24 respectively.
A dropout probability of 0.1 is used unless stated otherwise.
The number of counting class is set to 10, and the maximum number of spans is 8. 
The beam size is 3 by default, while the maximum amount of signed numbers $M$ is set to 4.
All texts are tokenized using WordPiece vocabulary~\cite{wu2016google}, and truncated to sequences no longer than 512 tokens.

\paragraph{Baselines}
Following the implementation of Augmented QANet (NAQANet)~\cite{dua2019drop}, we introduce a similar baseline called Augmented BERT (NABERT). 
The main difference is that we replace the encoder of QANet~\cite{yu2018fast} with the pre-trained Transformer blocks~\cite{devlin2018bert}.
Moreover, it also supports the prediction of various answer types such as span, arithmetic expression, and count number.

\begin{table}
	\begin{center}
		\begin{tabular}{l|cc}
			\toprule
			Model & EM & F1 \\ 
			\midrule
			MTMSN                & 76.7 & 80.5  \\
			w/o Q/P Vectors            & 75.1 &  79.2 \\
			w/o CLS Vector              & 74.0 & 78.4  \\
			Q/P Vectors Using Last Hidden          & 76.5 & 80.2  \\
			w/o Gated Span Prediction	     & 75.8 & 79.7  \\
			Combine Add/Sub with Negation  & 75.5 & 79.4  \\
			\bottomrule
		\end{tabular}
		\caption{\label{table:ablation2} Ablation tests of different architecture choices using \mtmsnlarge.}
	\end{center}
\end{table}

\subsection{Main Results}
Two metrics, namely Exact Match (EM) and F1 score, are utilized to evaluate models.
We use the official script to compute these scores. 
Since the test set is hidden, we only submit the best single model to obtain test results.

Table \ref{table:drop-result} shows the performance of our model and other competitive approaches on the development and test sets. 
MTMSN outperforms all existing approaches by a large margin, and creates new state-of-the-art results by achieving an EM score of 75.85 and a F1 score of 79.88 on the test set. 
Since our best model utilizes \bertlarge as encoder, we therefore compare \mtmsnlarge with the \nabertlarge baseline.
As we can see, our model obtains 12.07/13.19 absolute gain of EM/F1 over the baseline, demonstrating the effectiveness of our approach.
However, as the human achieves 95.98 F1 on the test set, our results suggest that there is still room for improvement.

\subsection{Ablation Study}
\begin{table}
	\begin{center}
		\small
		\begin{tabular}{l|ccccc}
			\toprule
			\multirow{2}*{ Type } & \multirow{2}*{ (\%) } & \multicolumn{2}{c}{ NABERT } & \multicolumn{2}{c}{ MTMSN  } \\
			&  & EM & F1 & EM & F1 \\ 
			\midrule
			Date				& 1.6  & 55.7 & 60.8 & 55.7 & 69.0 \\
			Number			    & 61.9 & 63.8 & 64.0 & 80.9 & 81.1 \\
			Single Span         & 31.7 & 75.9 & 80.6 & 77.5 & 82.8 \\
			Multi Span          & 4.8  & 0 & 22.7 & 25.1 & 62.8 \\
			\bottomrule
		\end{tabular}
		\caption{\label{table:gold-type} Performance breakdown of \nabertlarge and \mtmsnlarge by gold answer types.}
	\end{center}
\end{table}

\paragraph{Component ablation}
To analyze the effect of the proposed components, we conduct ablation studies on the development set. 
As illustrated in Table \ref{table:ablation1}, the use of addition and subtraction is extremely crucial: the EM/F1 performance of both the base and large models drop drastically by more than 20 points if it is removed.
Predicting count numbers is also an important component that contributes nearly 5\% gain on both metrics.
Moreover, enhancing the model with the negation type significantly increases the F1 by roughly 9 percent on both models. 
In brief, the above results show that multi-type answer prediction is vitally important for handling different forms of answers, especially in cases where discrete reasoning abilities are required.

We also report the performance after removing the multi-span extraction method.
The results reveal that it has a more negative impact on the F1 score.
We interpret this phenomenon as follows: producing multiple spans that are partially matched with ground-truth answers is much easier than generating an exactly-matched set of multiple answers. 
Hence for multi-span scenarios, the gain of our method on F1 is relatively easier to obtain than the one on EM.
Finally, to ablate arithmetic expression reranking, we simply use the arithmetic expression that has the maximum cumulative sign probability instead. 
We find that our reranking mechanism gives 1.8\% gain on both metrics for the large model. 
This confirms that validating expression candidates with their context information is beneficial for filtering out highly-confident but wrong predictions. 

\paragraph{Architecture ablation}
We further conduct a detailed ablation in Table \ref{table:ablation2} to evaluate our architecture designs.
First, we investigate the effects of some ``global vectors'' used in our model. 
Specifically, we find that removing the question and passage vectors from all involved computation leads to 1.3 \% drop on F1.
Ablating the representation of \texttt{[CLS]} token leads to even worse results.
We also try to use the last hidden representation (denoted as $\mathbf{M}_3$) to calculate question and passage vectors, but find that does not work.
Next, we remove the gating mechanism used during span prediction, and observe a nearly 0.8\% decline on both metrics.
Finally, we share parameters between the arithmetic expression component and the negation component, and find the performance drops by 1.1\% on F1.

\subsection{Analysis and Discussion}
\paragraph{Performance breakdown}
We now provide a quantitative analysis by showing performance breakdown on the development set.
Table \ref{table:gold-type} shows that our gains mainly come from the most frequent number type, which requires various types of symbolic, discrete reasoning operations. 
Moreover, significant improvements are also obtained in the multi-span category, where the F1 score increases by more than 40 points.
This result further proves the validity of our multi-span extraction method.

\begin{table}
\begin{center}
\small
\begin{tabular}{l|cccccc}
\toprule
\multirow{2}*{ Type } & \multicolumn{3}{c}{ NABERT } & \multicolumn{3}{c}{ MTMSN  } \\
  & (\%) & EM & F1 & (\%) & EM & F1 \\ 
\midrule
Span			   & 43.0 & 67.9 & 74.2 & 42.7 & 72.2 & 81.0 \\
Add/Sub			   & 43.6 & 62.0 & 62.1 & 32.4 & 78.1 & 78.2 \\
Count              & 13.4 & 62.4 & 62.4 & 13.4 & 70.4 & 70.4 \\
Negation           & 0 & 0 & 0 & 11.5 & 96.3 & 96.3 \\
\bottomrule
\end{tabular}
\caption{\label{table:pred-type} Performance breakdown of \nabertlarge and \mtmsnlarge by predicted answer types.}
\end{center}
\end{table}

We also give the performance statistics that are categorized according to the predicted answer types in Table \ref{table:pred-type}. 
As shown in the Table, the main improvements are due to the addition/subtraction and negation types.
We conjecture that there are two reasons for these improvements.
First, our proposed expression reranking mechanism helps validate candidate expressions.
Second, a new inductive bias that enables the model to perform logical negation has been introduced.
The impressive performance on the negation type confirms our judgement, and suggests that the model is able to find most of negation operations.
In addition, we also observe promising gains brought by the span and count types. 
We think the gains are mainly due to the multi-span extraction method as well as architecture designs.

\begin{figure}
	\center
	\includegraphics[width=.95\columnwidth]{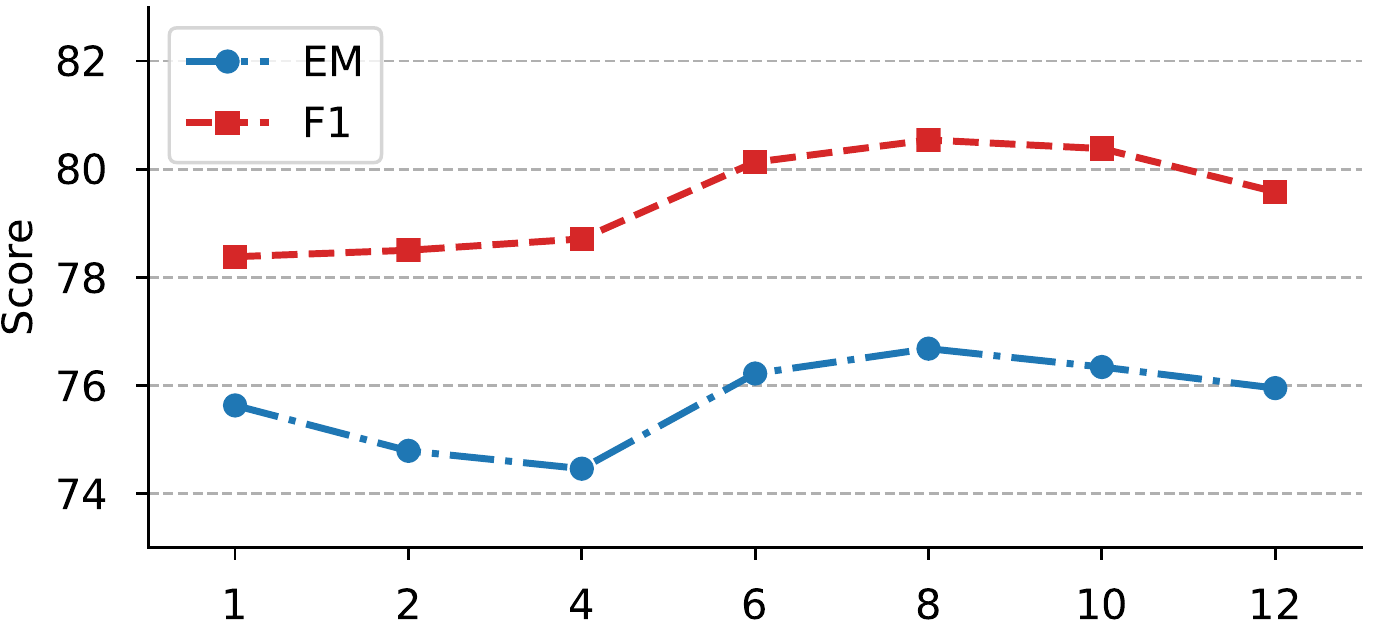}
	\caption{EM/F1 scores of \mtmsnlarge with different maximum numbers of spans.}
	\label{fig:max_num_spans}
\end{figure}

\paragraph{Effect of maximum number of spans}
To investigate the effect of maximum number of spans on multi-span extraction, we conduct an experiment on the dev set and show the curves in Figure \ref{fig:max_num_spans}.
We vary the value from 2 to 12, increased by 2, and also include the extreme value 1.
According to the Figure, the best results are obtained at 8. 
A higher value could potentially increase the answer recall but damage the precision by making more predictions, and a smaller value may force the model to produce limited number of answers, resulting in high precision but low recall.
Therefore, a value of 8 turns out to be a good trade-off between recall and precision.
Moreover, when the value decreases to 1, the multi-span extraction degrades to previous single-span scenario, and the performance drops significantly.

\paragraph{Effect of beam size and $M$}

\begin{figure}
	\center
	\includegraphics[width=.95\columnwidth]{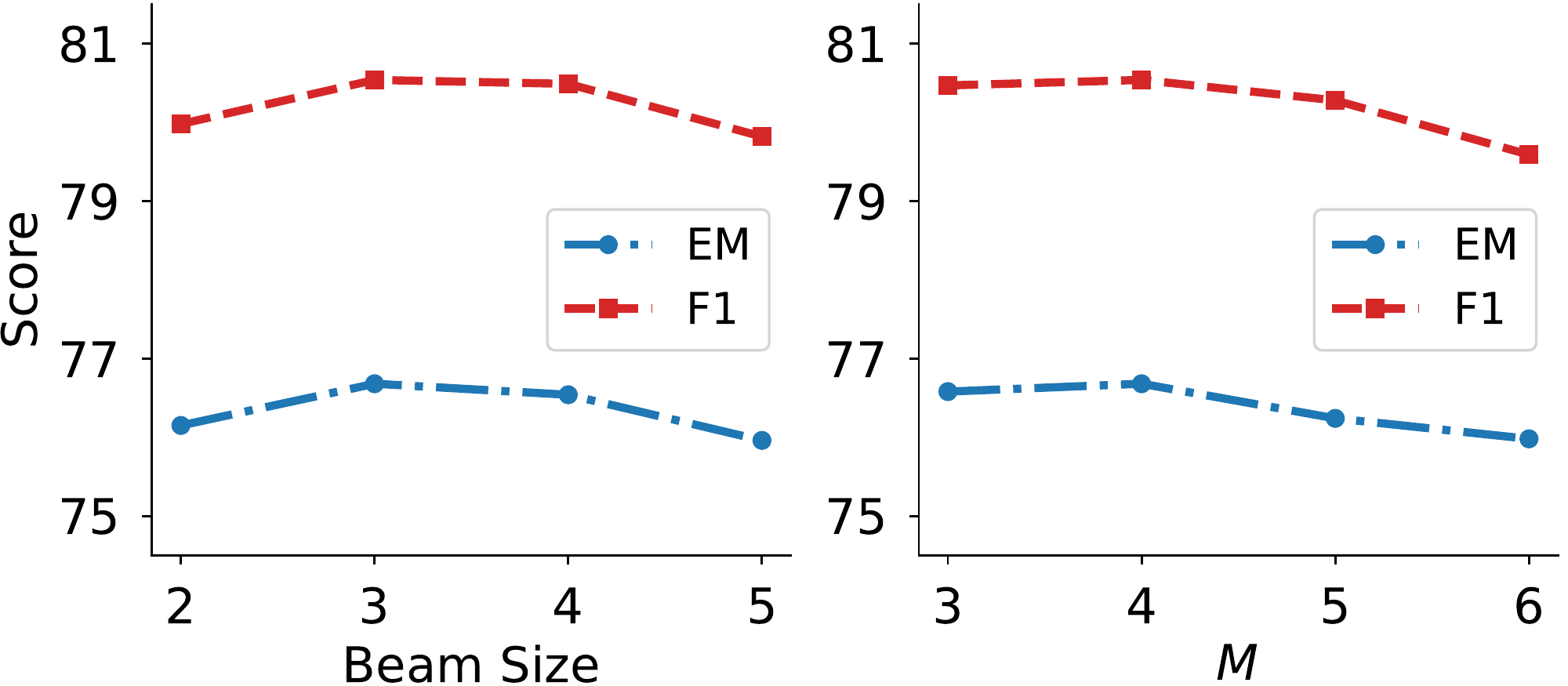}
	\caption{EM/F1 scores of \mtmsnlarge with different beam sizes and amounts of signed numbers ($M$).}
	\label{fig:beam_size_M}
\end{figure}
We further investigate the effect of beam size and maximum amount of signed numbers in Figure \ref{fig:beam_size_M}. 
As we can see, a beam size of 3 leads to the best performance, likely because a larger beam size might confuse the model as too many candidates are ranked, on the other hand, a small size could be not sufficient to cover the correct expression.
In addition, we find that the performance constantly decreases as the maximum threshold $M$ increases, suggesting that most of expressions only contain two or three signed numbers, and setting a larger threshold could bring in additional distractions.

\paragraph{Annotation statistics}
We list the annotation statistics on the DROP train set in Table \ref{table:statis}. 
As we can see, only annotating matching spans results in a labeled ratio of 56.4\%, indicating that DROP includes various answer types beyond text spans.
By further considering the arithmetic expression, the ratio increase sharply to 91.7\%, suggesting more than 35\% answers need to be inferred with numeral reasoning.
Continuing adding counting leads to a percentage of 94.4\%, and a final 97.9\% coverage is achieved by additionally taking negation into account.
More importantly, the F1 score constantly increases as more answer types are considered. This result is consistent with our observations in ablation study.

\begin{table}
	\begin{center}
		\small
		\begin{tabular}{l|ccccc}
			\toprule
			Configuration & Skipped & Kept & Ratio (\%) & F1  \\ 
			\midrule
			Span                & 33752 & 43657 & 56.4 & 38.9 \\
			+ $\clubsuit$            & 6384 & 71025 & 91.7 & 59.2 \\
			+ $\clubsuit$ + $\spadesuit$              & 4282 & 73127 & 94.4 & 63.6 \\
			+ $\clubsuit$ + $\spadesuit$ + $\heartsuit$           & 1595 & 75814 & 97.9 & 72.8 \\
			\bottomrule
		\end{tabular}
		\caption{\label{table:statis} Annotation statistics under different combinations of answer types in the DROP train set. ``Kept'' and ``Skipped'' mean the number of examples with or without annotation, respectively. $\clubsuit$ refers to Add/Sub, $\spadesuit$ denotes Count, and $\heartsuit$ indicates Negation. F1 scores are benchmarked using \mtmsnbase on the dev set.}
	\end{center}
\end{table}

\paragraph{Error analysis}
Finally, to better understand the remaining challenges, we randomly sample 100 incorrectly predicted examples based on EM and categorize them into 7 classes.
38\% of errors are incorrect arithmetic computations, 18\% require sorting over multiple entities, 13\% are due to mistakes on multi-span extraction, 10\% are single-span extraction problems, 8\% involve miscounting, another 8\% are wrong predictions on span number, the rest (5\%) are due to various reasons such as incorrect preprocessing, negation error, and so on.
See Appendix for some examples of the above error cases.

\section{Related Work}
\paragraph{Reading comprehension benchmarks}
Promising advancements have been made for reading comprehension due to the creation of many large datasets.
While early research used cloze-style tests~\cite{hermann2015teaching,Hill16}, most of recent works~\cite{Rajpurkar16,joshi2017triviaqa} are designed to extract answers from the passage.
Despite their success, these datasets only require shallow pattern matching and simple logical reasoning, thus being well solved~\cite{chen2016thorough,devlin2018bert}.
Recently, \citet{dua2019drop} released a new benchmark named DROP that demands discrete reasoning as well as deeper paragraph understanding to find the answers.
\citet{saxton2019analysing} introduced a dataset consisting of different types of mathematics problems to focuses on mathematical computation.
We choose to work on DROP to test both the numerical reasoning and linguistic comprehension abilities.

\paragraph{Neural reading models}
Previous neural reading models, such as BiDAF~\cite{seo2016bidirectional}, R-Net~\cite{wang2017gated}, QANet~\cite{yu2018fast}, Reinforced Mreader~\cite{hu2017reinforced}, are usually designed to extract a continuous span of text as the answer.
\citet{dua2019drop} enhanced prior single-type prediction to support various answer types such as span, count number, and addition/subtraction.
Different from these approaches, our model additionally supports a new negation type to increase answer coverage, and learns to dynamically extract one or multiple spans.
Morevoer, answer reranking has been well studied in several prior works~\cite{cui2016attention,wang2017evidence,wang2018multi,wang2018joint,hu2019retrieve}.
We follow this line of work, but propose ranking arithmetic expressions instead of candidate answers.

\paragraph{End-to-end symbolic reasoning}
Combining neural methods with symbolic reasoning was considered by \citet{graves2014neural,sukhbaatar2015end}, where neural networks augmented with external memory are trained to execute simple programs. 
Later works on program induction~\cite{reed2015neural,neelakantan2015neural,liang2016neural} extended this idea by using several built-in logic operations along with a key-value memory to learn different types of compositional programs such as addition or sorting.
In contrast to these works, MTMSN does not model various types of reasoning with a universal memory mechanism but instead deals each type with individual predicting strategies.

\paragraph{Visual question answering}
In computer vision community, the most similar work to our approach is Neural Module Networks~\cite{andreas2016neural}, where a dependency parser is used to lay out a neural network composed of several pre-defined modules.
Later, \citet{andreas2016learning} proposed dynamically choosing an optimal layout structure from a list of layout candidates that are produced by off-the-shelf parsers.
\citet{hu2017learning} introduced an end-to-end module network that learns to predict instance-specific network layouts without the aid of a parser.
Compared to these approaches, MTMSN has a static network layout that can not be changed during training and evaluation, where pre-defined ``modules'' are used to handle different types of answers. 

\section{Conclusion}
We introduce MTMSN, a multi-type multi-span network for reading comprehension that requires discrete reasoning over the content of paragraphs.
We enhance a multi-type answer predictor to support logical negation, propose a multi-span extraction method for producing multiple answers, and design an arithmetic expression reranking mechanism to further confirm the prediction.
Our model achieves 79.9 F1 on the DROP hidden test set, creating new state-of-the-art results.
As future work, we would like to consider handling additional types such as sorting or multiplication/division.
We also plan to explore more advanced methods for performing complex numerical reasoning.

 \section*{Acknowledgments}
 We would like to thank the anonymous reviewers for their thoughtful comments and insightful feedback.
 This work was supported by the National Key Research and Development Program of China (2016YFB100101).

\bibliography{sections/reference}
\bibliographystyle{acl_natbib}

\appendix

\section{Supplemental Material}
\begin{table*}[!t]
\centering
\footnotesize
\begin{tabular}{lcllcc} 
\toprule
Error Type & (\%) & Passage Snippets & Question & Answer & Prediction \\
 \midrule
 \midrule
\tabincell{c}{Arithmetic \\ computation} & 
38 &
\multicolumn{1}{m{4.5cm}}{\ldots{} there were 256,644 people living in Northern Cyprus. \ldots{} 2,482 born in the UK \ldots} & 
\multicolumn{1}{m{4.5cm}}{How many people in Northern Cyprus were not born in the UK?} & 
254,162 & 
144,923 \\
 \midrule
Sorting & 
18 &
\multicolumn{1}{m{4.5cm}}{\ldots{} 56.2\% were between the ages of 18 and 24; 16.1\% were from 25 to 44; 10.5\% were from 45 to 64 \ldots} & 
\multicolumn{1}{m{4.5cm}}{Which age group had the most people?} & 
\tabincell{c}{18 and \\ 24} & 
\tabincell{c}{from 45 \\ to 64} \\
 \midrule
\tabincell{c}{Multi-span \\ extraction} & 
13 &
\multicolumn{1}{m{4.5cm}}{Montana finished the regular season with \ldots{} 18 touchdowns, \ldots{} Craig was also a key contributor, \ldots{} and 10 touchdowns \ldots} & 
\multicolumn{1}{m{4.5cm}}{Which players scored ten or more touchdowns?} & 
\tabincell{c}{Craig, \\ Montana} & 
\tabincell{c}{Craig, \\ Rice} \\
 \midrule
\tabincell{c}{Single-span \\ extraction} & 
10 &
\multicolumn{1}{m{4.5cm}}{\ldots{} The sector decreased by 7.8 percent in 2002, before rebounding in 2003 with a 1.6 percent growth rate.} & 
\multicolumn{1}{m{4.5cm}}{In what year did the service sector have a sharp drop in growth rate?} & 
\tabincell{c}{2002} & 
\tabincell{c}{2003} \\
 \midrule
Counting & 
8 &
\multicolumn{1}{m{4.5cm}}{\ldots{} Macedonians with 338,358 inhabitants, Albanians with 103,891 \ldots{} Roma people with 23,475, Serbs (14,298 inhabitants) \ldots} & 
\multicolumn{1}{m{4.5cm}}{How many of the ethnic groups listed had more than 10,000 inhabitants in Skopje in 2002?} &
4 & 
3 \\
 \midrule
Span number & 
8 &
\multicolumn{1}{m{4.5cm}}{\ldots{} Tamil people form the majority of Chennais population. \ldots{} Chennais population was 80.7\% Hindu, 9.5\% Muslim \ldots} & 
\multicolumn{1}{m{4.5cm}}{Which religious people are the most common in Chennai?} & 
Hindu & 
\tabincell{c}{Hindu \\ Tamil} \\
\midrule
Other & 
5 &
\multicolumn{1}{m{4.5cm}}{\ldots{} 521,938 (6.6\%) of the economically active were unemployed \ldots} & 
\multicolumn{1}{m{4.5cm}}{How many in percent of the economically active were employed ?} & 
93.4 & 
48.4 \\
 \bottomrule
\end{tabular}
\caption{Error analysis on the DROP dev set. 100 randomly-sampled examples are classified into 7 categories. We list the error type, the ratio of each category, some representative examples, and our erroneous predictions.}
\label{tab:error_analysis}
\end{table*}

\end{document}